\documentclass[acmsmall,screen,nonacm]{acmart}

\usepackage{amsmath,amsfonts}
\usepackage[ruled,linesnumbered]{algorithm2e}

\SetCommentSty{mycommfont}
\usepackage{tikz}
\usetikzlibrary{arrows}
\usetikzlibrary{positioning}
\usepackage{mathtools}
\usepackage{listings}
\usepackage{pifont}
\usepackage{caption}
\usepackage{subcaption}
\usepackage{tcolorbox}
\usepackage{fontawesome}
\usepackage{mathdots}
\usepackage{multirow}
\usepackage{multicol}
\usepackage{threeparttable}
\usepackage{colortbl}
\usepackage{empheq}
\usepackage{varwidth}
\usepackage[frozencache=true,cachedir=./minted-cache]{minted}
\usepackage{makecell}
\usepackage{bbding}

\newcommand{\greyboxb}[2]{
\vspace{0.05cm}
    \begin{tcolorbox}[
        left=2pt, right=2pt, top=2pt, bottom=2pt,
        boxrule=0.2mm,
        leftrule=2mm,
        arc=0mm,
        colframe=black!40!white, 
        colback=black!5!white, 
        colbacktitle=black!50!white 
    ]
    \textbf{#1}{#2}
    \end{tcolorbox}
\vspace{0.05cm}
}
\begin{document}

\title{Verified Lifting of Deep learning Operators}

\author{Qi Zhan}
\orcid{0000-0002-6800-1857}
\affiliation{%
  \institution{The State Key Laboratory of Blockchain and Data Security, Zhejiang University}
   \city{Hangzhou}
  \country{China}
}
\email{qizhan@zju.edu.cn}

\author{Xing Hu}
\orcid{0000-0003-0093-3292}
\authornote{Corresponding Author}
\affiliation{
  \institution{The State Key Laboratory of Blockchain and Data Security, Zhejiang University}
   \city{Hangzhou}
  \country{China}
  }
\email{xinghu@zju.edu.cn}

\author{Xin Xia}
\orcid{0000-0002-6302-3256}
\affiliation{
  \institution{Zhejiang University}
  \city{Hangzhou}
   \country{China}
  }
\email{xin.xia@acm.org}

\author{Shanping Li}
\orcid{0000-0003-2615-9792}
\affiliation{
  \institution{The State Key Laboratory of Blockchain and Data Security, Zhejiang University}
   \city{Hangzhou}
  \country{China}
}
\email{shan@zju.edu.cn}


\begin{abstract}

Deep learning operators are fundamental components of modern deep learning frameworks. With the growing demand for customized operators, it has become increasingly common for developers to create their own.
However, designing and implementing operators is complex and error-prone, due to hardware-specific optimizations and the need for numerical stability.
There is a pressing need for tools that can summarize the functionality of both existing and user-defined operators.
To address this gap, this work introduces a novel framework for the verified lifting of deep learning operators, which synthesizes high-level mathematical formulas from low-level implementations. Our approach combines symbolic execution, syntax-guided synthesis, and SMT-based verification to produce readable and formally verified mathematical formulas. 
In synthesis, we employ a combination of top-down and bottom-up strategies to explore the vast search space efficiently;
In verification, we design invariant synthesis patterns and leverage SMT solvers to validate the correctness of the derived summaries;
In simplification, we use egraph-based techniques with custom rules to restore complex formulas to their natural, intuitive forms.
Evaluated on a dataset of deep learning operators implemented in Triton from the real world, our method demonstrates the effectiveness of synthesis and verification compared to existing techniques. This framework bridges the gap between low-level implementations and high-level abstractions, improving understanding and reliability in deep learning operator development.

\end{abstract}



\maketitle

\section{Introduction}

Deep learning (DL) is now widely applied in diverse fields, including computer vision~\cite{Bayoudh2021ASO}, natural language processing~\cite{9075398}, and software engineering~\cite{10.1145/3505243}.
As its adoption grows, the demands for performance and functionality in deep learning frameworks continue to increase.
The design and implementation of deep learning operators (DL operators) are critical for performance as they are building blocks of the DL framework.
DL operator is essentially an API call to perform tensor manipulations, that handle the numerical computations for model training and inference.
Conceptually, it can be viewed as a mathematical function that mappings input tensors to an output tensor, expressed as $Op: \text{list}[\text{tensor}] \rightarrow \text{tensor}$, where tensors are high-dimensional arrays.
For example, the softmax function is a widely used operator that converts the previous layer's output into a probability distribution.

The increasing complexity of DL models has amplified the need for customized operators.
Ideally, developers interact only with the interfaces provided by DL libraries without knowing individual DL operators.
GPU experts implement these operators to manage the utilization of hardware resources. 
However, as the complexity and demands of deep learning models continue to grow, developers increasingly need to (1) implement new DL operators or (2) optimize existing DL operators themselves. Many projects aim to accelerate both model inference and training by customizing DL operators. For example, Unsloth~\cite{unsloth}, FlagGems\footnote{https://github.com/FlagOpen/FlagGems}, and Liger-Kernel~\cite{hsu2024ligerkernelefficienttriton} implement custom RMSNorm~\cite{rmsnorm} and attention~\cite{dao2022flashattention} kernels. In addition, compiler frameworks such as Triton~\cite{triton}, which facilitate the productive development of custom kernels, are gaining popularity.

Implementing DL operators is challenging and error-prone, despite their mathematical simplicity.
To fully utilize the hardware resources, optimization techniques such as tiling (dividing computations into blocks)~\cite{tiling} and fusion (merging multiple operators into one)~\cite{kernel_fusion} are commonly employed, further increasing the complexity of implementation.
For example, highly optimized implementations like Flash attention~\cite{dao2022flashattention, dao2023flashattention2} are difficult to write and not easily understandable.
In addition, numerical stability is a critical concern~\cite{deepstability}, requiring developers to ensure the correctness of the implementations. Developers may transform the original mathematical definition of DL operators to improve numerical stability.
These complexities make understanding and implementing DL operators a significant challenge.

In summary, while implementing customized DL operators is becoming increasingly popular, it remains challenging.
Consequently, there is a pressing need for a summary of current implementation to assist developers.
We summarize the task as follows:


\begin{tcolorbox}[
    left=2pt, right=2pt, top=2pt, bottom=2pt,
    boxrule=0.2mm,
    leftrule=2mm,
    arc=0mm,
]
Given the source code of a DL operator, the objective is to derive its functionality as a mathematical formula $f$, such as $\text{softmax}(\mathbf{x})=\dfrac{\exp(\mathbf{x})}{\Sigma\exp(\mathbf{x})}$.
\end{tcolorbox}

Existing studies on the DL operator generally focus on (1) testing the implementation as a black box to find dimension or precision bugs~\cite{predoo} or (2) validating specific properties such as synchronization by treating it as concurrent programs~\cite{gpu1}.
However, solutions that verify the functionality of these operators are lacking. Such summaries could provide developers with a clear, formally verified understanding of an operator’s behavior, bridging the gap between low-level code and high-level mathematical intuition.
Developers can also input their highly optimized implementation and verify whether the generated formulas align with their expectations.

To address this gap, we propose a novel framework for verified lifting of DL operators. Our approach automatically synthesizes high-level mathematical formulas from low-level implementations, ensuring correctness and readability.
The framework comprises three key phases: synthesis, verification, and simplification.
First, we symbolically execute the program to extract the specification and search for a summary of the operator’s behavior.
The search space of the synthesis phase is huge, as the implementation of DL operators is usually complex and optimized.
To address this challenge, we explore top-down and bottom-up synthesis techniques to navigate the search space effectively.
Second, we formally verify the synthesized program to ensure the functionality of all valid inputs.
The complex mathematical computations in DL operators make it challenging for the solver to prove or disprove the formula.
To simplify the verification process, we design specific patterns that transform the invariant and postcondition into an equivalent but simpler formula, which is easier to verify.
Finally, we simplify the optimized program used for numerical stability to its natural mathematical form. It makes the program more readable and intuitive while preserving the equivalent.

Our implementation is built on the Triton~\cite{triton} and MLIR framework~\cite{mlir}.
In experiments, we collect 33 DL operator implementations from FlagGems and Triton.
Compared to our baseline, Tenspiler~\cite{qiu2024tenspiler}, our approach successfully synthesizes 32 operators and verifies 28 operators, whereas Tenspiler is limited to synthesizing only 13 operators.
We also convert the synthesized program into a PyTorch implementation and test its equivalence with the original operators using random inputs.
Among the successfully synthesized operators, our approach always achieves a shorter time cost than the baseline.
The experimental results show that our approach can lift the operators effectively.
In addition, we conducted an ablation study to analyze the contributions of different components of our approach. 
The methods we propose in the top-down synthesis methodology and pattern
designs in verification can improve the success rates of synthesis and verification by
88\% and 75\%, respectively.
The study shows the effectiveness of the proposed methods.

\textbf{Contributions.}
The main contributions of this paper can be summarized as follows:
\begin{itemize}
    \item To the best of our knowledge, this is the first work for verified lifting of deep learning operators.
    We present a framework capable of automatically lifting the implementation of deep learning operators to a high-level language.
    \item During the synthesis phase, we propose a novel approach to synthesize the program by divide and conquer; in the verification phase, we introduce invariant simplification and synthesis methods.
    \item We evaluated our approach on a set of benchmarks collected from real-world DL operators and demonstrated the effectiveness of our approach.
\end{itemize}

The remainder of this paper is organized as follows. In Section~\ref{sec:overview}, we demonstrate our approach using the softmax function as an example. Sections~\ref{sec:synthesis} and \ref{sec:verification} detail the synthesis and verification processes, respectively. The details of simplification and implementation are provided in Section~\ref{sec:impl}. Section~\ref{sec:evaluation} evaluates the effectiveness and practicality of our approach. Section~\ref{sec:discussion} discusses the limitations and threats to validity. Finally, we review related work in Section~\ref{sec:related-work} and conclude in Section~\ref{sec:conclusion}.

\section{Overview}\label{sec:overview}

\tikzset{
    >=stealth',
    punkt/.style={
           rectangle,
           rounded corners,
           draw=black, very thick,
           text width=6.5em,
           minimum height=2em,
           },
    pil/.style={
           ->,
           very thick,
           shorten <=2pt,
           shorten >=2pt,}
}

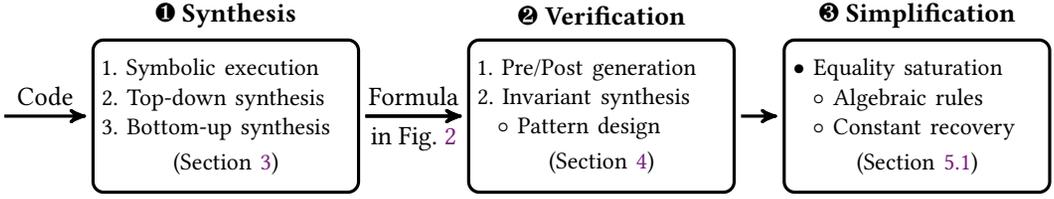
\begin{figure*}[t]
\centering
\begin{tikzpicture}

 \node[punkt, minimum height=2cm, minimum width=3.5cm, text width=3.3cm] (synthesis) at (4, 8) {
    \small{1. Symbolic execution \\ 2. Top-down synthesis \\ 3. Bottom-up synthesis \\ \begin{center} (Section~\ref{sec:synthesis}) \end{center}}
 };
 \node at ([yshift=0.3cm]synthesis.north) {\ding{182}~\textbf{Synthesis}};

 \node[punkt, minimum height=2cm, minimum width=3.5cm, text width=3.3cm] (verify) at ([xshift=5cm]synthesis) {
    \small{1. Pre/Post generation \\ 2. Invariant synthesis \\ \phantom{1} $\circ$ Pattern design \\
    \begin{center}
        (Section~\ref{sec:verification})
    \end{center}}
 };
\node at ([yshift=0.3cm]verify.north) {\ding{183}~\textbf{Verification}};

 \node[punkt, minimum height=2cm, minimum width=3.5cm, text width=3.3cm] (simplification) at ([xshift=4.2cm]verify) {
    \small{$\bullet$ Equality saturation \\ \phantom{1} $\circ$ Algebraic rules \\ \phantom{1} $\circ$ Constant recovery \\ \begin{center} (Section~\ref{sec:simplification}) \end{center}
    }
 };
 \node at ([yshift=0.3cm]simplification.north) {\ding{184}~\textbf{Simplification}};

\draw[pil] ([xshift=-35pt, yshift=0pt]synthesis.west) -- node[above, align=center] {Code} ([yshift=0pt]synthesis.west);
\draw[pil] ([yshift=0pt]synthesis.east) -- node[above, align=center] {Formula} ([yshift=0pt]verify.west);
\draw[pil] ([yshift=0pt]synthesis.east) -- node[below, align=center] {in Fig.~\ref{fig:syntax}} ([yshift=0pt]verify.west);
\draw[pil] ([yshift=0pt]verify.east) -- node[above, align=center] {} ([yshift=0pt]simplification.west);


\end{tikzpicture}
\caption{Overview of our approach}
\label{fig:approach}
\end{figure*}

The overall framework of our approach is illustrated in Figure~\ref{fig:approach}.
Given the source code of a specific operator, the framework outputs a mathematical summary of its behavior.
The lifting process is divided into three phases: 
\ding{182}~Synthesis: A mathematical formula summarizing the implementation is generated. Given the source code of a DL operator, we begin with symbolically executing it to extract a specification. We then employ top-down and bottom-up synthesis methods to construct a program that satisfies the specification.
\ding{183}~Verification: The correctness of the synthesized formula is validated through the generation of preconditions, postconditions, and loop invariants. During the verification process, we identify a pattern that simplifies the procedure and enables the SMT solver to make faster decisions.
\ding{184}~Simplification: The synthesized formula is simplified by equality saturation. We apply arithmetic equivalence and constant recovery rules to restore the formula to its natural mathematical form.

We demonstrate how our approach lifts the softmax function. 
The code below implements the softmax function in Triton~\cite{triton}. It consists of two parts: host code and kernel code. The host code controls the program, delegating computational tasks and launching kernels across multiple threads. 
In contrast, the kernel code performs the actual computations on the GPU.
In each thread, the kernel code first loads data (Line 13) using the row index corresponding to the thread index \verb|program_id|, performs the concrete softmax computation (Lines 14–17), and then stores the results in the output tensor (Line 20).

\begin{minted}[linenos,xleftmargin=20pt,breaklines]{python}
def softmax(x):
    BLOCK_SZIE = 4 # compile-time constant, assume to be 4
    grid, _ = x.shape
    y = torch.empty_like(x)
    softmax_kernel[grid](y, x, stride_o, stride_i, cols, BLOCK_SIZE)
    return y

@triton.jit
def softmax_kernel(y, x, stride_o, stride_i, num_cols, BLOCK_SIZE):
    row_index = program_id
    row_start_ptr = x + row_index * stride_i
    x_pointers = row_start_ptr + arange(0, BLOCK_SIZE)
    row = load(x_pointers)
    safe_row = row - max(row, axis=0)
    numerator = exp(safe_row)
    denominator = sum(numerator, axis=0)
    out = numerator / denominator
    y_row_start_ptr = y + row_index * stride_o
    y_pointers = y_row_start_ptr + col_offsets
    store(y_pointers, out)
\end{minted}

\ding{182} First, we symbolically execute the kernel code using a small tensor size on each thread. Assume the input tensor $\mathbf{x}$ is $ [x_1, x_2, x_3]$\footnote{we use \textbf{boldface} to represent a tensor and italics to represent a scalar.}, the corresponding output tensor is $\textbf{y} = [y_1, y_2, y_3]$, and we can get

\begin{equation}\label{eq:softmax_pointwise}
y_i = \frac{\exp(x_i - \max(x_1, x_2, x_3))}{\exp(x_1 - \max(x_1, x_2, x_3)) + \exp(x_2 - \max(x_1, x_2, x_3)) + \exp(x_3 - \max(x_1, x_2, x_3))}, i= 1, 2, 3.
\end{equation}

The next step is synthesizing a tensor program equivalent to the pointwise formula in Equation (\ref{eq:softmax_pointwise}).
The key idea is a divide-and-conquer approach: partition the output tensor into subtensors based on structural similarity and then recursively synthesize a formula for each subtensor.
Using the synthesis algorithm, we derive the formula:
\begin{equation}\label{eq:softmax}
f(\mathbf{x}) = \frac{\exp(\mathbf{x} - \max(\mathbf{x}))}{\Sigma\exp(\mathbf{x} - \max(\mathbf{x}))}.
\end{equation}
To check the candidate formula matches the output, we employ an SMT solver to check:
$$\lnot (f(x_1) = y_1 \land f(x_2) = y_2 \land f(x_3) = y_3).$$
The solver returns an “unsat” result, indicating that the function is correct.

\ding{183} Thus far, we have only demonstrated that the formula holds for tensors of a \textbf{bounded} length (3 in this case). However, it is necessary to prove the full functional equivalence for tensors of arbitrary size.
Extensive research exists on verifying the correctness of code in \textbf{unbounded} cases.
In our approach, we employ classical Hoare logic~\cite{hoare}, a widely used method for program verification.
In Hoare logic, a Hoare triple is expressed as $\{P\} S \{Q\}$, where $P$ is the precondition, $S$ is the program or statement being verified, and $Q$ is the postcondition that must hold after executing $S$, assuming $P$ holds beforehand.
The precondition $P$ in this problem is $\forall x, \exp(x) > 0$, which depicts mathematical properties for function $\exp$ to prevent division by zero, and the postcondition $Q$ is derived directly from the synthesized formula:

$$
\forall i, 0\le i < \text{length} \implies y[i] = \frac{\exp(x_i - m)}{\sum_{j=0}^3 \exp(x[i //4*4 + j]-m)}, \text{where}~ m = \max_{j=0}^3(x[i//4*4+j])
$$

The tensor $\mathbf{x}$  can be viewed as a matrix with dimensions $n \times 4$, each row representing a block of four elements. The summation and maximum operations are performed row-wise, and the expression $i // 4 * 4$ is used to identify the starting index of the corresponding row.

In the verification phase, we consider the thread launch sequentially. The host code can be viewed as a single loop:

\begin{minted}[xleftmargin=20pt,breaklines,frame=single]{python}
program_id = 0
while program_id < length / 4:
    softmax_kernel(y, x, program_id, ...)
    program_id += 1
\end{minted}

Since the program includes a loop over the thread index, a loop invariant is required to verify the postcondition. The condition $C$ for loop is $\text{pid} < \text{length} / 4$.
We introduce a series of techniques to generate the loop invariant for operators and speed up the invariant synthesis.
Consequently, the loop invariant for the softmax function is:
\begin{equation*}
I \triangleq \forall i, 0 \le i < 4*\text{pid} \implies y[i] = \frac{\exp(x_i - m)}{\sum_{j=0}^3 \exp(x[i //4*4 + j]-m)},
\end{equation*}

where the definition of $m$ is the same as above.
By encoding the precondition, postcondition, condition, body, and loop invariant into the SMT solver, we construct the following formula:
$$
F \triangleq (P \implies I) \land (I \land C \implies I') \land  (\lnot C \land I \implies Q).
$$
While the SMT solver determines that the formula $\lnot F$ is unsatisfiable, the correctness of the program is verified.

\ding{184} The program derived in Equation (\ref{eq:softmax}) does not exactly match the ideal mathematical result.
This is because the expression $\exp(x)$ may overflow when $x$ is a large floating-point number.
To improve numerical stability, developers often use the trick $(\mathbf{x} - \max(\mathbf{x}))$ in real-world implementations.
This adjustment ensures that the intermediate results do not overflow and the expression remains equivalent to the original softmax function, as dividing the numerator and denominator by the same value does not change the result.

To enhance understanding of the program, the developer may wish to simplify the formula to reveal its underlying structure. We achieve this by simplifying the program by rewriting rules. The rules include general arithmetic properties for basic operators and mathematical functions.
We outline the step-by-step simplification process of the formula in Equation (\ref{eq:softmax}) as follows. The formula at each step is listed in the left column, with the corresponding rules presented in the right column.

\begin{table}[h!]
\centering
\begin{tabular}{ll}
\textbf{Formula}                                                                                   & \textbf{Rewrite Rule}                                 \\ \hline
$\exp(\mathbf{x} - \max(\mathbf{x})) / \operatorname{sum}(\exp(\mathbf{x} - \max(\mathbf{x})))$     & $\exp(a-b) = \exp(a) / \exp(b)$                      \\ 
$= \exp(\mathbf{x} - \max(\mathbf{x})) / (\operatorname{sum}(\exp(\mathbf{x}) / \exp(\max(\mathbf{x}))))$ & $\operatorname{sum}(a / b) = \operatorname{sum}(a) / b$ \\ 
$= \exp(\mathbf{x} - \max(\mathbf{x})) / \operatorname{sum}(\exp(\mathbf{x})) / \exp(\max(\mathbf{x}))$ & $\exp(a-b) = \exp(a) / \exp(b)$                      \\ 
$= \exp(\mathbf{x}) / \exp(\max(\mathbf{x})) / \operatorname{sum}(\exp(\mathbf{x})) / \exp(\max(\mathbf{x}))$ & $(a/b) / (c/b) = a/c$                                \\ 
$= \exp(\mathbf{x}) / \operatorname{sum}(\exp(\mathbf{x}))$                                         &                                                   \\ 
\end{tabular}
\label{tab:softmax_rewrite}
\end{table}

As a result, the formula after simplification is $\frac{\exp(\mathbf{x})}{\Sigma\exp(\mathbf{x})}$, a familiar softmax function in textbook.
In summary, we have demonstrated lifting low-level operator code to a high-level mathematical formula, ensuring both correctness and readability through synthesis, verification, and simplification.
While the example focuses on synthesizing a single output tensor from a single input tensor, our approach can synthesize multiple outputs from multiple inputs.

\section{Synthesis Primer}\label{sec:synthesis}

Synthesis aims to generate a program that satisfies a given specification within a search space.
We formulate our problem as a classical syntax-guided synthesis (SyGuS) problem, as proposed by~\cite{sygus},
which consists of three key components:

\begin{figure}
\begin{empheq}[box=\fbox]{align*}
        \text{Expr}~\textit{e}   & \coloneq e~\text{binop}~e \mid \text{unop } e \mid \text{const}~c \mid \text{input tensor}~i \mid \text{function}(e) \\
                                 & \mid \max(e) \mid \text{sum}(e) \mid \text{permute}~(e) \\
                                 & \mid \text{if}~e~\text{then}~e~\text{else}~e \\
        \text{unop }             & \coloneq - \mid \neg \mid \cdots                            \\
        \text{binop }            & \coloneq + \mid - \mid \times \mid \div \mid \odot \mid \cdots \\
        \text{function }         & \coloneq \exp \mid \log \mid \sin \mid \cdots
\end{empheq}
\caption{Language Syntax}
\label{fig:syntax}
\end{figure}

\begin{enumerate}
    \item \textbf{Specification} defines the properties that the synthesized tensor program should satisfy. In our case, the specification is not provided by the user but is automatically generated through symbolic evaluation, as detailed in Section~\ref{sec:symbolic}.
    \item \textbf{Search Space} represents the space of potential solutions, defined by a context-free grammar presented in Figure~\ref{fig:syntax}. The grammar is designed to be intuitive and includes basic arithmetic operations and mathematical functions on input tensors and constant values.
    \item \textbf{Search Algorithm} searches the candidate programs that satisfy the specification, as described in Algorithm~\ref{alg:synthesis}. To address the huge space, we employ a top-down synthesis approach that recursively decomposes the operator into smaller subproblems. and use a bottom-up strategy to handle the remaining cases.
\end{enumerate}

\SetAlgoNoEnd
\SetKwComment{Comment}{$\triangleright$ }{}
\begin{algorithm}
    \KwIn{inputs, output, maxDepth}
    \SetKwFunction{FMain}{synthesis}
    \SetKwFunction{FSub}{bottom-up}
    \SetKwProg{Fn}{Function}{:}{}
    \Fn{\FMain{inputs, output}}{
     {$\triangleright$ \scriptsize consider all binary operations such as $+, -, \times, \div$}\;
        \For{op in allBinaryOp()}{ \label{line:binary} 
       
            \If{left, right $\gets$ splitBy(output, op)}{  
                left $\gets$ synthesis(inputs, left)\;
                right $\gets$ synthesis(inputs, right)\;
                \If{(left op right)(inputs) = output} {
                     \Return{left op right}
                }
            }
        }  
         {$\triangleright$ \scriptsize consider all functions such as $\exp, \sin, \tanh, \text{sqrt}$}\;
        \For{op in MathFunctions()}{ \label{line:math}
            \If{arg $\gets$ splitBy(output, op)}{
                arg $\gets$ synthesis(inputs, arg)\;
                \Return{op arg}\;
            }
        }
         {$\triangleright$ \scriptsize consider sum, matrix multiplication}\;
        \If{guessSum(output)}{ \label{line:sum}
            args $\gets$ synthesis(splitBy(output, sum))\;
            \If{(sum(args))(inputs) = output} {
                \Return{sum(args)}
            }
        }
        \Return{bottom-up(inputs, output, depth)} \label{line:resort}
    }

    \Fn{\FSub{inputs, output, maxDepth}}{
        plist $\gets$ set of all terminals\;
        depth $\gets$ 1\;
        \While{depth $\le$ maxDepth}{
        plist $\gets$ grow(plist)\;
        plist $\gets$ elimTypeMismatch(plist, inputs)\;
        plist $\gets$ elimValueeMismatch(plist, inputs)\;
        \For{p in plist}{
            \If{p(inputs) = output}{
                \Return{p}
            }
        }
        depth $\gets$ depth + 1\;
        }
    }
    \caption{Synthesis Algorithm}
    \label{alg:synthesis}
\end{algorithm}

In the following sections, we detail the process of obtaining the specification and designing the search algorithm.

\subsection{Symbolic Evaluation}\label{sec:symbolic}

The first step in our synthesis process is to determine the specification of the implementation.
A common approach involves selecting random inputs from the numerical domain, executing the program, and using (inputs, output) pairs as examples for program synthesis. This reduces the problem to a classical program-by-example task~\cite{pbe}.
 However, this approach fails to leverage the rich information embedded in the source code fully. A more significant issue is that different implementations of DL operators which represent the same mathematical formula, may not produce identical outputs for the same input. For example, Flash Attention~\cite{dao2023flashattention2} and the standard attention implementation in PyTorch~\cite{pytorch} may introduce a 1\% relative error for some inputs.

In this work, we employ a different approach. We interpret the whole program as symbolic execution and use symbolic value in the output tensor as specifications. The whole process is as follows:
\begin{enumerate}
    \item Setup. We begin by initializing the state and fixing the shape of each input tensor, represented as $\mathbf{input}_1, \dots, \mathbf{input}_n$, where $n$ is the number of parameters of the function. All elements in the input tensors are assigned symbolic values. Each thread is assigned a corresponding thread index, and the program is executed sequentially.
    \item Execution. Ordinary arithmetic operations are modeled as their symbolic counterparts, while mathematical functions such as $\exp$, $\log$, and $\sin$ are represented as uninterpreted functions in first-order logic. 
    Complex operations such as $\max(x_1, x_2, x_3)$ are encoded explicitly for the SMT solver as ``$\text{if}~x_1 > (\text{if}~x_2 > x_3~\text{then}~x_2~\text{else}~x_3)~\text{then}~x_1~\text{else}~(\text{if}~x_2 > x_3~\text{then}~x_2~\text{else}~x_3)$''.
    \item Result collection. Since the shape of the input tensor is fixed, the program terminates after a finite number of steps, regardless of the values inside the array are symbolic. We collect the symbolic expression of the output tensor after termination, represented as $\mathbf{output}$.
\end{enumerate}

The symbolic result for each element of the output tensor serves as a constraint for the objective function. The specification that the synthesized function  $f$  must satisfy is determined by combining all point-wise constraints:
$$
f(\mathbf{input}_1, \dots, \mathbf{input}_n) = \mathbf{output}.
$$


\subsection{Top-down Search}

The top-down search serves as the primary technique to synthesize the programs. 
The intuition behind the design is the rich structure information from symbolic evaluation provides the opportunity to synthesize the program in a \textit{divide-and-conquer} manner.
Consider Equation~\ref{eq:softmax_pointwise} as an example, each element in the output tensor is a division of two terms, which can be decomposed into two subproblems.
\begin{equation*}
    \begin{cases}
        y_{1i} = \exp(x_i - \max(x_1, x_2, x_3)), i = 1,2,3 \\
        y_{2i} = \exp(x_1 - \max(x_1, x_2, x_3)) + \exp(x_2 - \max(x_1, x_2, x_3)) + \exp(x_3 - \max(x_1, x_2, x_3))
    \end{cases} 
\end{equation*}

Assuming the subproblems $y_{1i}, y_{2i}$ have been successfully synthesized and are represented as $f_1$  and $f_2$, the final result is expressed as $\frac{f_1}{f_2}$.
In general, we analyze the structure of each element in the output tensor, hypothesize the possible subproblems, and synthesize each of them recursively. If the synthesis is successful, we can reconstruct the original program by combining the solution of subproblems with the corresponding operations.

Returning to the algorithm, we handle binary operations in Line~\ref{line:binary}. If the output can be partitioned by the binary operation, we synthesize the left and right subtensors and combine them using the operation. The handling of mathematical functions is similar, as shown in Line~\ref{line:math}. In addition, we attempt to decompose the elements through summation and infer whether the program can be constructed using summation or matrix multiplication, as shown in Line~\ref{line:sum}.
If all elements in a row are formulated as $ a_1 + a_2 + \dots + a_n $, we consider it as $\sum \text{synthesis}(\mathbf{a})$. Similarly, if all elements are formulated as $a_1b_1 + a_2b_2 + \dots + a_nb_n$, we consider it as $\mathbf{a}\mathbf{b}$.

\subsection{Bottom-up Search}

When the top-down search fails, we resort to the bottom-up enumeration.
Bottom-up synthesis is a standard technique in program synthesis~\cite{AlbarghouthiGK13}.
The idea is to explicitly construct all possible programs from terminals in the language.
Terminals in our problems consist of input tensors as well as their transpose and all constants existing in the program.
While the search space is too large, we prune the search space by the following rules:

\begin{itemize}
    \item \textbf{Eliminate type-mismatch programs.} We eliminate programs whose dimensions do not match the output tensor.
   Ill-typed programs are also eliminated. For example, matrix multiplication of $a$ and $b$ is ill-typed if the second dimension of $a$ does not match the first dimension of $b$.
    \item \textbf{Eliminate value-mismatch programs.} We eliminate programs that cannot possibly be part of the solution.
    For example, if the first element in the output tensor contains $a[0], b[0]$, and the current subprogram contains $a[1], c[0]$.
    This subprogram cannot contribute to the solution since $c[0]$ should never appear in the output tensor.
    In such cases, we can confidently eliminate the program as it does not satisfy the specification.
\end{itemize}

Once we generate a candidate program, we consider $\lor_{i}~f(\text{inputs}_i) \ne \text{output}_i$ and check its satisfiability using an off-the-shelf SMT solver. If the solver returns an ``unsat'' result, it indicates that the program satisfies the specification.

To summarize, we derive the specification by symbolically executing the program with a small input size. The synthesis approach combines top-down and bottom-up methods to explore the solution space efficiently. Finally, an SMT solver is employed to verify the correctness of the synthesis results.

\section{Loop Verification}\label{sec:verification}

Given a candidate program synthesized from a \textbf{bounded} set of program states, we must verify whether it is correct for \textbf{unbounded} states.
There is extensive literature on how to verify that a block of code is valid to a given pre and post-conditions. 
A well-established method for achieving this is through the construction of verification conditions, which implies that the code is valid to its pre- and postconditions.
Verification conditions are typically constructed using Hoare logic~\cite{hoare}, a widely used framework for program verification.
By the classical Hoare triple $\{P\} S\{Q\}$, where $P$ is precondition, $S$ is the program, and $Q$ is postcondition.
The most challenging part is handling loops within the program, as a loop invariant $I$ is required for verification.
To prove the postcondition in a loop, we need to prove three statements:
\begin{align}
    &\forall s. P(s) \implies I(s) \\
    &\forall s. I(s) \land C(s) \implies I(\text{body}(s)) \\
    &\forall s. I(s) \land \neg C(s) \implies Q(s)
\end{align}

The first statement (3) ensures that the loop invariant is initially true based on the precondition. The second statement (4) guarantees that if the loop invariant holds at the beginning of an iteration and the loop condition is satisfied, the invariant will continue to hold after executing the loop body. This step confirms that the invariant is maintained throughout the loop.
The third statement (5) asserts that if the invariant holds but the loop condition is not satisfied, then the postcondition must hold.
By combining the three statements, we can prove that the postcondition holds for the program.
In the following sections, we detail the process of generating precondition, postcondition, and potential invariants.

\subsection{Precondition and Postcondition Generation}

The precondition $P$ in our problem consists of constraints applied to both uninterpreted functions and the arguments. We manually define specific properties as preconditions during verification for mathematical functions modeled by first-order logic, which are challenging to define precisely. For example, we assume that $\forall x, \exp(x) > 0$, which ensures $\sum \exp(\mathbf{x}) > 0$. This guarantees that the summation remains positive, thereby preventing division by zero errors in the softmax kernel. Regarding arguments, we assume that certain variables are positive in specific operators. For example, in RMSNorm~\cite{rmsnorm}, the condition $\epsilon > 0$ is essential to prove $\sum \mathbf{x}^2 + \epsilon > 0$, which similarly avoids division by zero.

The postcondition $Q$ specifies the final property that the program must satisfy after execution. We need to transform the high-level tensor program into a point-wise format.
To achieve it, we iterate the tensor program recursively with the index variable $i$ and generate the formula by indexing the array by $i$.
The process is easy for most constructors in our grammar. 
For example, addition $\mathbf{c} = \mathbf{a} + \mathbf{b} $ is transformed to $ \forall i. c[i] = a[i] + b[i]$.
However, it is nontrivial for constructors involving column operations such as $\max$ or $\sum$.
Due to the constraints of writing kernel code, we must assign a specific column number rather than an arbitrary length for operations like max and sum. For instance, if \verb|BLOCK_SIZE| is 4, the summation operation $\sum x$ for a matrix $x$ will be instantiated as $x[i] + x[i+1] + x[i+2] + x[i+3]$, where $i = 4m, m \in \mathbb{N}$. In this case, we iterate through the tensor program using the index variables $i, i+1, i+2,$ and $i+3$ separately and combine the results according to the operation.

\subsection{Invariant Generation}

Invariant generation is a crucial step in verifying programs~\cite{Bradley2007}.
As the pre and post-conditions are automatically derived, the challenging part lies in the construction of the invariant.
Invariant synthesis is naturally formulated as a $\exists.\forall$ problem, which seeks to find a predicate \textbf{a} predicate that hold true for \textbf{all} states.
In this work, we focus on the generation of array manipulation invariants,
which is a long history and hard problem~\cite{array1, array2}.
In array manipulation, the invariants themselves also involve universal quantification, e.g. 
$$
\forall~0 \le i < n, 0 \le j < m \implies a[i][j] = 2 * b[i][j] + 1,
$$
which makes it harder to find a satisfactory assignment for an SMT solver~\cite{arrayinv}.
We do not claim to solve the invariant generation problem about array manipulation in general.
Instead, we study the specific patterns that are common in the programs we are interested in and propose a solution to them.

\faLightbulbO~\textbf{Insight:} Many operator designs follow a similar pattern: different threads perform the same task, which aligns with the postconditions. As each thread completes its task, some indices in the output arrays satisfy the final condition.
\begin{minted}[xleftmargin=20pt,breaklines, frame=single]{c}
for i in range(0, program_id):
    for j in range(0, n):
        index = i * n + j
        b[index] = f(a[index], ..., a[index + BLOCK_SIZE - 1])
\end{minted}

The effect of the function $f$ is exactly the postcondition.
In this case, we can abstract the effects of each thread and consider it as a property $P$.
The invariants can be formally specified by a special universally quantified formula $\forall i, 0 \le i < n \implies P(i)$, where $P(i)$ is the point-wise postcondition we want to prove. By doing so, we prove an equivalent formula rather than the original, more complex one.
We use tensor addition $\mathbf{z} = \mathbf{x} + \mathbf{y}$ as a simple example.
The postcondition $\land_{k=0}^3 z[4i+k] = x[4i+k] + y[4i+k]$ can be transformed to $P(4i) \land P(4i+1) \land P(4i+2) \land P(4i+3)$.

In our implementation, we extract all possible indices from the transition body and substitute the universal quantifier with these indices. If we can prove that the substituted formula is equivalent to the original formula for a specific index $k$, we can safely replace it with $P(k)$.
After simplification, we transform the problem of verifying array manipulation into verifying index manipulation, which is simpler.
For the index problem, we employ a template-based approach to generate the index component of the loop invariant using a linear function, such as $\forall i, 0 \leq i < ax + b \implies P(i)$.
The SMT solver determines valid assignments for $a$ and $b$ that satisfy the three verification properties. 
We provide the constructed formula to an off-the-shelf SMT solver to check its satisfiability.

For programs that do not meet the pattern, we use the basic template-based invariant synthesis, as mentioned above.
While we focus on a single loop in the program, the process can be easily extended to handle nested loops.

\section{Simplification and Implementation}\label{sec:impl} 

\subsection{E-graphs based Simplification}\label{sec:simplification}

As mentioned above, the complexity of mathematical computations in DL operators often results in highly optimized yet opaque formulas.
For example, the computation may introduce intermediate transformations, such as subtracting the maximum value in the softmax function, to ensure numerical stability.
These formulas can be hard to understand because they are different from the usual way in the textbook.
Simplifying these formulas back to their natural form is crucial for facilitating developer understanding. 
To address these, we employ an e-graphs-based simplification process.

An e-graph is a data structure that stores an equivalence relation over terms.
The core idea is to rewrite expressions iteratively based on the provided rules while constructing and updating an e-graph. 
In our use case, this process involves applying specific transformation rules to simplify complex expressions into human-readable and verifiable forms.
By equality saturation, we ensure that the simplification process explores all potential rewrites and identifies the most optimized or canonical form of the expression.
We design three sets of rules:

\begin{enumerate}
    \item General Arithmetic Rules. These rules reflect fundamental properties of arithmetic operations, such as commutative, associative, distributive, and cancellation laws. Our implementation follows prior work that seeks to improve the accuracy of floating-point computations by leveraging these properties~\cite{egg}.
    \item Specialized rules for mathematical functions. These set of rules is designed to handle mathematical functions like $\exp$, $\log$, and $\sin$. They are based on the intrinsic properties of these functions, such as $\exp(a - b) = \frac{\exp(a)}{\exp(b)}$ or $\log(a \cdot b) = \log(a) + \log(b)$. These transformations simplify expressions while preserving their numerical equivalence.
    \item Constant recovery rules. Constants in the program, such as $\log_2 e$, are replaced by floating-point numbers like $1.4426950216293335$. Our approach aims to recover the original mathematical constant from the floating-point representation, making the formula more meaningful and easier for developers to understand.
\end{enumerate}

\subsection{Implementation}

Our implementation is built on the Triton~\cite{triton} and MLIR~\cite{mlir} frameworks. Triton offers a Python-based language and an MLIR-based compiler for writing highly efficient custom DL operators, which has gained popularity in the deep learning community. Our analysis uses Triton IR, an MLIR dialect generated by the Triton compiler as inputs. We customize Triton's source code to support more interfaces as Python bindings.
While the main synthesis and verification algorithms are detailed in the previous sections, the following discusses the additional implementation components not covered earlier.

\subsubsection{Symbolic Evaluation} 

We implement the symbolic evaluation as an interpreter in Python, using the NumPy library and the Z3 solver~\cite{z3}. All tensors are represented as arrays in NumPy, while their values are modeled as real numbers in Z3 rather than floating-point numbers. As our primary focus is on functionality verification, we choose Z3’s real theory to reduce the solver's efforts. To support operations on symbolic arrays, we utilized NumPy’s universal function mechanism, which extends standard array operations to symbolic arrays. The operations we implemented for one or two symbolic values can be seamlessly extended to arrays of any dimension, significantly reducing the manual effort required for implementing symbolic array operations.

\subsubsection{Annotation}

As the host and kernel codes are split natively, certain annotations are required to facilitate symbolic execution. The type of each argument in the kernel function must be annotated to distinguish array pointers and constant values. All constants used within the kernel, as well as the total number of threads to be created, must also be annotated. For arrays, tensor shapes need to be explicitly annotated, as kernel code treats all tensors as one-dimensional arrays. While it is theoretically possible to infer shape information from constraints extracted from the host code, this approach often lacks sufficient information and may produce inaccurate results. Given this trade-off, we choose to annotate the necessary information manually.

\subsubsection{Egglog based egraph}

We use egglog, a tool for equality saturation based on e-graphs, and Datalog for simplification~\cite{egglog}. 
We use egglog’s Python binding\footnote{\url{https://egglog-python.readthedocs.io/latest/}} to integrate the simplification process into our framework. We design a total of 26 rules, which cover common patterns in DL operator expressions. These rules not only improve numerical stability but also restore the expressions to a form that aligns with their natural mathematical representation, facilitating verification and developer understanding.


\section{Evaluation}\label{sec:evaluation}

\begin{table*}
    \centering
    \caption{Overall results}
    \label{RQ1}
    \begin{tabular}{@{}lcccccc@{}}
        \toprule
        \multirow{2}{*}{Kernel} & \multicolumn{2}{c}{{{Tenspiler}}} & \multicolumn{3}{c}{\textbf{\emph{Our Approach}}}      & \multirow{2}{*}{Formula}                \\ \cmidrule(l){2-3}  \cmidrule(l){4-6}  & S$^*$ + V$^*$                                      & Time   & S$^*$ & V$^*$ & Time   \\ \midrule
        Binary   & \Checkmark      & 3s & \Checkmark     & \Checkmark  & 0.8s & $\mathbf{y} = \mathbf{x}_1~\text{op}~ \mathbf{x}_2 , \text{op} \in \{+, -,\times, \div\} $ \\  
        Cat    & \Checkmark     & 3s & \Checkmark     & \Checkmark  & 0.8s  & $\mathbf{y} = \mathbf{x}  $  \\  
        Neg    & \Checkmark     & 3s & \Checkmark     & \Checkmark  & 0.8s  & $\mathbf{y} = -\mathbf{x}  $  \\
        Reciprocal    & \Checkmark     & 3s & \Checkmark     & \Checkmark  & 1s & $\mathbf{y} = 1/{\mathbf{x}}  $  \\
        Zeros    & \Checkmark     & 3s & \Checkmark     & \Checkmark  & 0.8s & $\mathbf{y} = 0$  \\
        Exp  & \Checkmark    & 2s & \Checkmark     & \Checkmark  & 0.9s & $\mathbf{y} = \exp\mathbf{x}$  \\ 
        Sin, Cos & \XSolidBrush    & - & \Checkmark     & \Checkmark  & 0.8s & $\mathbf{y} = \text{op}~\mathbf{x}, \text{op} \in\{ \sin ,\cos\}$   \\ 
        Log & \XSolidBrush    & - & \Checkmark     & \Checkmark  & 0.9s & $\mathbf{y} = \log\mathbf{x}$  \\
        Abs & \XSolidBrush    & - & \Checkmark     & \Checkmark  & 0.8s  & $\mathbf{y} = |\mathbf{x}|$  \\
        RSqrt    & \Checkmark     & 3s & \Checkmark     & \Checkmark  & 0.9s & $\mathbf{y} = 1/{\sqrt{\mathbf{x}}}$  \\ 
        Tanh    & \XSolidBrush      & 3s & \Checkmark     & \Checkmark  & 0.9s & $\mathbf{y} = \tanh{\mathbf{x}}$ \\
        Sum    & \Checkmark     & 2s & \Checkmark     & \Checkmark  & 1s & $\mathbf{y} = \sum\mathbf{x}$\\ 
        Max    & \Checkmark     & 7s & \Checkmark     & \Checkmark  & 1s & $\mathbf{y} = \max\mathbf{x}$\\ 
        MatMul & \XSolidBrush      & - & \Checkmark     & \XSolidBrush  & 3s & $\mathbf{y}= \mathbf{x}_1  \mathbf{x}_2$ \\ 
        \midrule
        Relu    & \XSolidBrush      & - & \Checkmark     & \Checkmark  & 0.9s & $\mathbf{y} = \begin{cases}\mathbf{x}& \text{if }\mathbf{x}> 0, \\ 0 & \text{if }\mathbf{x}\leq 0. \end{cases} $ \\ 
        LeakyRelu    & \XSolidBrush      & - & \Checkmark     & \Checkmark  & 0.9s & $\mathbf{y} = \begin{cases} 0.01\mathbf{x} & \text{if }\mathbf{x}> 0, \\ 0 & \text{if }\mathbf{x}\leq 0. \end{cases} $ \\ 
        SquareRelu & \XSolidBrush      & - & \Checkmark     & \Checkmark  & 0.9s & $\mathbf{y} = \begin{cases} \mathbf{x}^2 & \text{if }\mathbf{x}> 0, \\ 0 & \text{if }\mathbf{x}\leq 0. \end{cases} $ \\ 
        Gelu    & \XSolidBrush      & - & \Checkmark     & \Checkmark  & 1s &  $\mathbf{y}= \frac12\mathbf{x}(1 + \tanh(\sqrt{\frac{2}{\pi}} \mathbf{x}(1 + A^* \mathbf{x}^2) ))$ \\
        SwiGLU   & \XSolidBrush      & - & \Checkmark     & \Checkmark  & 1s & $\mathbf{y}= \mathbf{x}_1 \mathbf{x}_2 /{(1 + e^{-\mathbf{x}})}$  \\ 
        Sigmoid & \XSolidBrush      & - & \Checkmark     & \Checkmark & 0.8s & $\mathbf{y}= 1/{(1 + e^{-\mathbf{x}})}$ \\ 
        Silu & \XSolidBrush      & - & \Checkmark     & \Checkmark & 1s & $\mathbf{y}= \mathbf{x}/{(1 + e^{-\mathbf{x}})} $\\ 
        \midrule
        Softmax1    & \Checkmark     & 13s & \Checkmark     & \Checkmark  & 3s & \multirow{2}{*}{$ \mathbf{y}= \dfrac{e^{\mathbf{x}}}{\sum e^{\mathbf{x}}}$}  \\ 
        Softmax2    & \XSolidBrush      & - & \Checkmark     & \XSolidBrush  & 3s &   \\ 
        LogSoftmax    & \XSolidBrush      & - & \Checkmark     & \Checkmark & 2s & $ \mathbf{y}= \log({e^{\mathbf{x}}}/{\sum e^{\mathbf{x}})}$ \\
        Layernorm   & \XSolidBrush      & - & \XSolidBrush      & \XSolidBrush  & - &  $\mathbf{y}= \dfrac{\mathbf{x} - \mu}{\sqrt{\sigma^2 + \epsilon}} \odot \gamma + \beta$\\ 
        RMSnorm   & \XSolidBrush      & - & \Checkmark     & \XSolidBrush  & 14m & $\mathbf{y}= \dfrac{\mathbf{x}}{\sqrt{\sigma^2 + \epsilon}} \odot \gamma$ \\ 
        GeluMul & \XSolidBrush      & - & \Checkmark     & \Checkmark & 1s & $\mathbf{y}= \frac12 \mathbf{x}_1 \mathbf{x}_2 (1 + \tanh(\sqrt{\frac{2}{\pi}} \mathbf{x}_1  (1 + A \mathbf{x}_1^2)))$ \\ 
        SiluMul   & \XSolidBrush      & - & \Checkmark     & \Checkmark & 0.9s & $\mathbf{y}= \mathbf{x}_1 \odot \mathbf{x}_2/({1 + e^{-\mathbf{x}_1}}) $ \\ 
        Attention & \XSolidBrush      & - & \Checkmark     & \XSolidBrush  & 5s & $\mathbf{y}= \text{softmax}(\mathbf{Q}\mathbf{K}^T)\mathbf{V}$\\ 
        \midrule
        \textbf{All}  & 13      &  & \textbf{32}     & \textbf{28} &  \\
        \bottomrule
    \end{tabular}
        \begin{tablenotes}
        \footnotesize
        \item * The symbol \XSolidBrush~signifies failure, lack of support or timeout, and \Checkmark~denotes successful synthesis or verification.
        \item * S stands for Synthesis, V stands for Verification.
        \item * $A = 0.044715$
    \end{tablenotes}
\end{table*}

To evaluate the effectiveness of our approach, we conducted experiments following three research questions.

\begin{enumerate}
\renewcommand{\labelenumi}{\textbf{RQ.\theenumi}}
    \item How effective and efficient are our approach and methods for synthesizing and verifying DL operators?
    \item Can our proposed top-down synthesis, bottom-down prune and verification techniques enhance the effectiveness of our approach?
    \item Does the simplification improve the formula?
\end{enumerate}

\subsection{Experimental Setup}

\subsubsection{Dataset Collection}\label{sec:dataset}

We collect the dataset from Triton tutorials\footnote{https://triton-lang.org/main/getting-started/tutorials/index.html}, FlagGems\footnote{https://github.com/FlagOpen/FlagGems}, Liger-Kernel~\cite{hsu2024ligerkernelefficienttriton} and modify the source code to align with our framework.
The collected kernels can be categorized into three groups: (1) Activation Functions such as Relu~\cite{Agarap2018DeepLU}, (2) Fused kernels such as flash attention~\cite{dao2022flashattention}, and (3) Foundation kernels, such as tiling matrix multiplication.
DL operators involving logical operations, indexing, masking, embedding, tensor shape manipulations, or random operations are excluded, as mathematical formulas cannot easily describe their behavior and are beyond the scope of this work.
For certain DL operators, we collect multiple implementations for comparison.
It is important to note that DL operators are more complex than they may initially appear, even for simple operations such as addition. The host code divides these operations into smaller tasks, then processed by multiple threads.
As a result, we collected a total of 33 operators.
We compile the collected source code to Triton IR as our input.
All \verb|BLOCK_SIZE| is set to be $4$ in the experiments.

\subsubsection{Baseline}

As our approach is the first work on lifting DL operators, there is no directly comparable baseline in the exact same area. Therefore, we select Tenspiler~\cite{qiu2024tenspiler} as our baseline, a tool designed to lift low-level tensor programs to high-level code rather than focus on DL operators. Tenspiler also utilizes synthesis and verification techniques to perform the entire lifting process. Since DL operators can be regarded as specialized tensor programs, a direct comparison between our approach and Tenspiler is possible, particularly in terms of the effectiveness of synthesis and verification.

Tenspiler cannot directly process DL operators written in Triton. To address this, we first identify equivalent operators in Tenspiler's dataset and use their corresponding implementations. For operators without an equivalent, we manually implement the required C++ code. To ensure functional correctness, we replace kernel calls with equivalent for-loop implementations. We set the timeout to an hour, following its original configuration.

We conducted all experiments on machines with an Intel CPU at 2.9GhZ with access to 64GB of main memory and Ubuntu 18.04 (64 bit) as the operating system.
\subsection{Experimental Results}

\subsubsection{RQ1: Comparison with Baselines}

To evaluate the effectiveness of our approach, we conducted experiments on the dataset. We compared our approach to Tenspiler and the overall results are shown in Table~\ref{RQ1}. For each kernel, we list the mathematical formula and report the success or failure of synthesis and verification, along with the corresponding time (if successful). Since synthesis and verification are coupled in Tenspiler, we report their success rate together. 
Our framework successfully synthesizes and verifies 28 operators out of the 33 tested, while Tenspiler was able to synthesize only 13 operators.
To further ensure correctness, we transform the mathematical formula into Python programs using PyTorch for all successfully synthesized operators. We then use randomized tests to verify the equivalence between the Triton implementation and the PyTorch version.
These results demonstrate the effectiveness of our approach for DL kernel lifting.
Our detailed findings are as follows:
\begin{enumerate}
    \item Tenspiler and our approach perform well on simple DL operators, such as basic binary operations, as they are relatively straightforward.
However, Tenspiler does not support many built-in operators commonly used in more complex DL tasks, which fail synthesis. For activation kernels involving complex mathematical computations and floating-point numbers, our approach also outperforms the baseline.
    \item Among the successfully synthesized operators, our approach always achieves a shorter time cost than Tenspiler. This improvement can be attributed to combining our top-down synthesis strategy and pattern designs for verification. Using these techniques, we reduce the search space complexity during synthesis and streamline the verification process for invariants.
    \item Our approach fails to synthesize the LayerNorm operator within the given time, as recovering results such as $x - \mu$ proves challenging for our top-down synthesis strategy. Among the operators that fail verification, the invariants for MatMul and Flash Attention are too complex, exceeding the scope of our current framework.
\end{enumerate}

Another advantage of our approach is the decoupling of synthesis and verification. While our method may not verify results for complex DL operator implementations, such as Flash Attention, users can still gain insights from the synthesized results, even if they remain unverified. In contrast, Tenspiler cannot provide any information on programs that are difficult to verify.

\greyboxb{Answer to RQ1:}{ Compared to Tenspiler, our approach successfully synthesizes 19 more kernels and verifies 15 more kernels, which shows the effectiveness of our approach. }

\subsubsection{RQ2: Ablation Study}

To evaluate the contribution of top-down synthesis and pattern in verification in the effectiveness of our approach, we performed an ablation study. 
We create the following variants:

\begin{itemize}
    \item \textsc{-Top}: We removed the top-down synthesis strategy in the synthesis phase and used only bottom-up enumeration to generate the program.
    \item \textsc{-Bot}: We removed the pruning technique in the synthesis and only used brute-force enumeration.
    \item \textsc{-Verify}: We removed the special pattern used to simplify the verification condition. Instead, we directly use template-based synthesis for the invariant generation.
\end{itemize}

We evaluated these variants using the same dataset described in Section~\ref{sec:dataset}. 
The results are shown in Table~\ref{RQ2}.
We list the successfully synthesized kernels for the first two variants and the successfully verified kernels for the last variant.

\begin{itemize}
    \item \textsc{-Top}: Removing the top-down search strategy results in 15 fewer successfully synthesized operators. It shows the critical role of top-down synthesis in improving the accuracy of the synthesis process. Without it, the synthesis becomes less structured, leading to a larger and more challenging search space. The classical bottom-up enumeration approach can only synthesize simple operators, such as addition, within the given time constraints.
    \item \textsc{-Bot}: Removing the pruning technique in bottom-up synthesis results in two fewer successful operators within the given timeout. The synthesis time for FlashAttention increases from 5 seconds to 10 minutes.
    \item \textsc{-Verify}: Removing the verification pattern results in 15 fewer successful cases. Our program involves numerous mathematical computations that are challenging for the SMT solver to verify. In our experiment, Z3 returns ``unknown'' even for simple cases such as $a+b$. This demonstrates the importance of replacing the complex invariant with a general property.
\end{itemize}

\begin{table}[h]
    \centering
    \caption{Ablation Study: Synthesis and Verification}
    \begin{subtable}[t]{0.48\textwidth}
        \centering
        \caption{Synthesis}
        \begin{tabular}{lcc}
            \toprule
            Approach           & Synthesis \\ \midrule
            Top + Bot & 32        \\
            \textsc{-Top}      & 17 ($\downarrow$15)  \\
            \textsc{-Bot}      & 30 ($\downarrow$2)  \\
            \bottomrule
        \end{tabular}
        \label{synthesis_ablation}
    \end{subtable}
    \hfill
    \begin{subtable}[t]{0.48\textwidth}
        \centering
        \caption{Verification}
        \begin{tabular}{lcc}
            \toprule
            Approach           & Verification\\ \midrule
            Verify & 28                 \\
            \textsc{-Verify}   & 16 ($\downarrow$12) \\ 
            \bottomrule
        \end{tabular}
        \label{verification_ablation}
    \end{subtable}
    \label{RQ2}
\end{table}

\greyboxb{Answer to RQ2:}{ Our approach benefits from the top-down synthesis methodology and pattern designs in verification, which can improve the success rates of synthesis and verification by 88\% and 75\%, respectively. }

\subsubsection{RQ3: Simplification}

In the last RQ, we evaluate how much our simplification process can improve the formula. 
Based on the experimental results, we conclude the situation that the simplification can cover:

\begin{itemize}
    \item Numeric stability. As demonstrated in the softmax example, developers often modify formula computations while preserving semantic equivalence. In such cases, our approach can successfully recover the original formula, as observed with softmax, log-softmax, and attention in the dataset.
    \item Constant recovery. Our method recovered the constant value $0.01$ in the LeakyRelu operators. This is because our dataset has few constants in the DL operators. The approach is expected to yield more substantial benefits for datasets with more DL operators containing constants.  
    \item Another interesting thing is that the simplification can bring some special improvements to the formula. For example, in the Gelu and GeluMul kernels, the $\tanh$ function is expanded and the overall formula is simplified.
\end{itemize}

\greyboxb{Answer to RQ3:}{ The simplification process improved the readability of 7 DL operators in the dataset.  }

\section{Discussion}\label{sec:discussion}

\subsection{Limitations}

While our approach can synthesize many DL operators successfully, the current implementation has several limitations.

\begin{enumerate}
    \item Manual Annotations. Our approach relies on detailed manual annotations to synthesize operator implementations. This requirement increases the manual effort, making the process less scalable, especially for large or complex operators. Moreover, Annotation errors may propagate through the synthesis and verification stages, potentially affecting the results' correctness.
    \item Restricted Scope. As mentioned in the dataset section~\ref{sec:dataset}, operators that cannot be easily described using mathematical formulas, such as masking are excluded. This limits the scope of our approach to a subset of operators that focus on numerical computations.
    \item Focus on Forward Kernels. The current implementation only targets forward kernels which is sufficient for model inference. However, it leaves the backward passes, which are critical for training, unaddressed.
    \item Dependency on SMT solver. The results depend on the performance and correctness of the Z3 solver. Failures or “unknown” results from the solver can negatively impact the framework’s effectiveness.
\end{enumerate}






\subsection{Threats to Validity}

\subsubsection{Internal Validity}
Synthesizing and verifying the implementation of DL operators rely on detailed annotations, which require manual effort and may introduce human errors. The correctness of preconditions and invariants used in synthesis and verification depends heavily on the accuracy of these annotations.
In our framework, all numbers are treated as ``Real'' due to the limited support for ``Float'' numbers in SMT solvers. In this regard, we follow the baseline design proposed in Tenspiler~\cite{qiu2024tenspiler}. While it simplifies the verification process, it may obscure bugs introduced by floating-point arithmetic.
The loop invariants generated during the verification phase rely on specific patterns. The framework might fail to construct valid invariants for operators that deviate from these patterns, limiting its effectiveness.

\subsubsection{External Validity}

The evaluated operators are primarily drawn from real-world examples in Triton, FlagGems, and related repositories, but they may not fully capture the diversity of DL operator implementations. 
In addition, our dataset consists of DL operators written using the Triton framework. Implementing DL operators in Triton is easier than in CUDA, as Triton automatically handles complexities such as memory coalescing and shared memory management. 
However, this simplification may limit the ability to represent the full range of complexities and variations encountered in other implementations, such as those written in CUDA. 

\section{Related Work}\label{sec:related-work}

\subsection{Verified Lifting}

Our approach can be classified in the category of verified lifting. The concept of verified lifting was first introduced by Cheng et al.\cite{querysynthesis} to synthesize SQL queries from imperative code. They start with a low-level implementation and synthesize a high-level specification that behaves identically to the implementation. \cite{stencil} demonstrates a novel combination of program synthesis and verification to lift stencil computations from low-level Fortran code to a high-level summary expressed using a predicate language. The use of symbolic evaluation for synthesis in our approach is inspired by bounded symbolic execution in the STUN compiler\cite{stencil}.

As our baseline, Tenspiler~\cite{qiu2024tenspiler} is the closest work in the context of lifting tensor programs. Based on metalift framework~\cite{metalift}, they also utilize synthesis and verification to lift tensor programs from low-level code to high-level code. However, we differ from Tenspiler in the following ways:
\begin{enumerate}
\item The grammar in Tenspiler is quite different from ours. While Tenspiler only considers summation, max operations, and basic arithmetic operations, we consider a wider range of operations commonly used in deep learning operators.
\item The overall synthesis process in our approach is more comprehensive and tailored to the specific needs of deep learning operator lifting.
\end{enumerate}

Another interesting work is TF-Coder~\cite{tf-coder}, which synthesizes TensorFlow programs from input-output examples. TF-Coder combines a bottom-up weighted search with a series of value- and type-based pruning techniques to accelerate the search process. In contrast, our approach does not rely on concrete input-output examples; instead, we use symbolic execution to synthesize a symbolic formula. This difference enables a more flexible and efficient design of the synthesis algorithm.

\subsection{Verify and test CUDA Programs}

Deep learning operators are often implemented in CUDA to leverage the parallelism of GPU.
Several works focus on verifying the properties of CUDA programs.
While they mainly focus on the verification of the concurrency and synchronization of the CUDA programs, our work focuses on the functionalities.
\cite{gpu1} first use an SMT-based approach to verify GPU kernel functions by encoding the thread interleave. It can detect data races or incorrect barriers. \cite{gpu2} and \cite{gpu3} consider the producer-consumer synchronization and data races in GPU programs, and propose a method to analyze and verify it, separately.

Another line of work that focuses on validating deep learning operators.
Predoo~\cite{predoo} is the first work to focus on precision testing of deep learning operators. It treats the testing of DL operators as a search problem, aiming to maximize output precision errors. By doing so, it successfully triggers larger precision errors compared to the declared error threshold.
Duo~\cite{duo} and Chen~\cite{metamorphic} use differential testing and metamorphic testing, respectively, to evaluate deep learning operators.
Duo focuses on comparing the outputs of deep learning models by testing the same input with different model implementations, detecting inconsistency in their behavior to identify potential bugs. Chen, on the other hand, use specific properties or relations that should hold true across different inputs. 
Compared to these works, our approach considers a completely different way to assure the correctness of the deep learning operators.

\section{Conclusion}\label{sec:conclusion}

In this paper, we present a novel framework for the verified lifting of deep learning operators, bridging the gap between low-level implementations and high-level mathematical representations. Our approach combines symbolic execution, syntax-guided synthesis, and formal verification to extract and verify mathematical formulas from highly optimized DL operator implementations.
Our framework improves developer understanding and provides a reliable method for verifying the correctness of operator implementations.
We compare our approach against Tenspiler, a state-of-the-art tool for lifting tensor programs. The results demonstrate that our approach outperforms Tenspiler in both success rate and speed.
This work opens up new opportunities for enhancing developer productivity and ensuring the correctness of DL operator implementations.


\bibliographystyle{ACM-Reference-Format}
\bibliography{ref}


\begin{thebibliography}{37}


\ifx \showCODEN    \undefined \def \showCODEN     #1{\unskip}     \fi
\ifx \showDOI      \undefined \def \showDOI       #1{#1}\fi
\ifx \showISBNx    \undefined \def \showISBNx     #1{\unskip}     \fi
\ifx \showISBNxiii \undefined \def \showISBNxiii  #1{\unskip}     \fi
\ifx \showISSN     \undefined \def \showISSN      #1{\unskip}     \fi
\ifx \showLCCN     \undefined \def \showLCCN      #1{\unskip}     \fi
\ifx \shownote     \undefined \def \shownote      #1{#1}          \fi
\ifx \showarticletitle \undefined \def \showarticletitle #1{#1}   \fi
\ifx \showURL      \undefined \def \showURL       {\relax}        \fi
\providecommand\bibfield[2]{#2}
\providecommand\bibinfo[2]{#2}
\providecommand\natexlab[1]{#1}
\providecommand\showeprint[2][]{arXiv:#2}

\bibitem[Bra(2007)]%
        {Bradley2007}
 \bibinfo{year}{2007}\natexlab{}.
\newblock \bibinfo{booktitle}{\emph{Invariant Generation}}.
\newblock \bibinfo{publisher}{Springer Berlin Heidelberg}, \bibinfo{address}{Berlin, Heidelberg}, \bibinfo{pages}{311--346}.
\newblock
\showISBNx{978-3-540-74113-8}
\urldef\tempurl%
\url{https://doi.org/10.1007/978-3-540-74113-8_12}
\showDOI{\tempurl}


\bibitem[Agarap(2018)]%
        {Agarap2018DeepLU}
\bibfield{author}{\bibinfo{person}{Abien~Fred Agarap}.} \bibinfo{year}{2018}\natexlab{}.
\newblock \showarticletitle{Deep Learning using Rectified Linear Units (ReLU)}.
\newblock \bibinfo{journal}{\emph{ArXiv}}  \bibinfo{volume}{abs/1803.08375} (\bibinfo{year}{2018}).
\newblock
\urldef\tempurl%
\url{https://api.semanticscholar.org/CorpusID:4090379}
\showURL{%
\tempurl}


\bibitem[Albarghouthi et~al\mbox{.}(2013)]%
        {AlbarghouthiGK13}
\bibfield{author}{\bibinfo{person}{Aws Albarghouthi}, \bibinfo{person}{Sumit Gulwani}, {and} \bibinfo{person}{Zachary Kincaid}.} \bibinfo{year}{2013}\natexlab{}.
\newblock \showarticletitle{Recursive Program Synthesis}. In \bibinfo{booktitle}{\emph{Computer Aided Verification - 25th International Conference, {CAV} 2013, Saint Petersburg, Russia, July 13-19, 2013. Proceedings}}. \bibinfo{publisher}{Springer}, \bibinfo{pages}{934--950}.
\newblock
\urldef\tempurl%
\url{https://doi.org/10.1007/978-3-642-39799-8\_67}
\showDOI{\tempurl}


\bibitem[Alur et~al\mbox{.}(2013)]%
        {sygus}
\bibfield{author}{\bibinfo{person}{Rajeev Alur}, \bibinfo{person}{Rastislav Bodik}, \bibinfo{person}{Garvit Juniwal}, \bibinfo{person}{Milo M.~K. Martin}, \bibinfo{person}{Mukund Raghothaman}, \bibinfo{person}{Sanjit~A. Seshia}, \bibinfo{person}{Rishabh Singh}, \bibinfo{person}{Armando Solar-Lezama}, \bibinfo{person}{Emina Torlak}, {and} \bibinfo{person}{Abhishek Udupa}.} \bibinfo{year}{2013}\natexlab{}.
\newblock \showarticletitle{Syntax-guided synthesis}. In \bibinfo{booktitle}{\emph{2013 Formal Methods in Computer-Aided Design}}. \bibinfo{pages}{1--8}.
\newblock
\urldef\tempurl%
\url{https://doi.org/10.1109/FMCAD.2013.6679385}
\showDOI{\tempurl}


\bibitem[Bayoudh et~al\mbox{.}(2021)]%
        {Bayoudh2021ASO}
\bibfield{author}{\bibinfo{person}{Khaled Bayoudh}, \bibinfo{person}{Raja Knani}, \bibinfo{person}{Fayçal Hamdaoui}, {and} \bibinfo{person}{Abdellatif Mtibaa}.} \bibinfo{year}{2021}\natexlab{}.
\newblock \showarticletitle{A survey on deep multimodal learning for computer vision: advances, trends, applications, and datasets}.
\newblock \bibinfo{journal}{\emph{The Visual Computer}}  \bibinfo{volume}{38} (\bibinfo{year}{2021}), \bibinfo{pages}{2939 -- 2970}.
\newblock
\urldef\tempurl%
\url{https://api.semanticscholar.org/CorpusID:235410640}
\showURL{%
\tempurl}


\bibitem[Bhatia et~al\mbox{.}(2023)]%
        {metalift}
\bibfield{author}{\bibinfo{person}{Sahil Bhatia}, \bibinfo{person}{Sumer Kohli}, \bibinfo{person}{Sanjit~A Seshia}, {and} \bibinfo{person}{Alvin Cheung}.} \bibinfo{year}{2023}\natexlab{}.
\newblock \showarticletitle{Building Code Transpilers for Domain-Specific Languages Using Program Synthesis (Experience Paper)}. In \bibinfo{booktitle}{\emph{37th European Conference on Object-Oriented Programming (ECOOP 2023)}}. Schloss Dagstuhl--Leibniz-Zentrum f{\"u}r Informatik.
\newblock


\bibitem[Chen et~al\mbox{.}(2024)]%
        {metamorphic}
\bibfield{author}{\bibinfo{person}{Jinyin Chen}, \bibinfo{person}{Chengyu Jia}, \bibinfo{person}{Yunjie Yan}, \bibinfo{person}{Jie Ge}, \bibinfo{person}{Haibin Zheng}, {and} \bibinfo{person}{Yao Cheng}.} \bibinfo{year}{2024}\natexlab{}.
\newblock \showarticletitle{A Miss Is as Good as A Mile: Metamorphic Testing for Deep Learning Operators}.
\newblock \bibinfo{journal}{\emph{Proc. ACM Softw. Eng.}} \bibinfo{volume}{1}, \bibinfo{number}{FSE}, Article \bibinfo{articleno}{89} (\bibinfo{date}{July} \bibinfo{year}{2024}), \bibinfo{numpages}{23}~pages.
\newblock
\urldef\tempurl%
\url{https://doi.org/10.1145/3660796}
\showDOI{\tempurl}


\bibitem[Cheung et~al\mbox{.}(2013)]%
        {querysynthesis}
\bibfield{author}{\bibinfo{person}{Alvin Cheung}, \bibinfo{person}{Armando Solar-Lezama}, {and} \bibinfo{person}{Samuel Madden}.} \bibinfo{year}{2013}\natexlab{}.
\newblock \showarticletitle{Optimizing database-backed applications with query synthesis}. In \bibinfo{booktitle}{\emph{Proceedings of the 34th ACM SIGPLAN Conference on Programming Language Design and Implementation}} (Seattle, Washington, USA) \emph{(\bibinfo{series}{PLDI '13})}. \bibinfo{publisher}{Association for Computing Machinery}, \bibinfo{address}{New York, NY, USA}, \bibinfo{pages}{3–14}.
\newblock
\showISBNx{9781450320146}
\urldef\tempurl%
\url{https://doi.org/10.1145/2491956.2462180}
\showDOI{\tempurl}


\bibitem[Daniel~Han and team(2023)]%
        {unsloth}
\bibfield{author}{\bibinfo{person}{Michael~Han Daniel~Han} {and} \bibinfo{person}{Unsloth team}.} \bibinfo{year}{2023}\natexlab{}.
\newblock \bibinfo{booktitle}{\emph{Unsloth}}.
\newblock
\urldef\tempurl%
\url{http://github.com/unslothai/unsloth}
\showURL{%
\tempurl}


\bibitem[Dao(2024)]%
        {dao2023flashattention2}
\bibfield{author}{\bibinfo{person}{Tri Dao}.} \bibinfo{year}{2024}\natexlab{}.
\newblock \showarticletitle{Flash{A}ttention-2: Faster Attention with Better Parallelism and Work Partitioning}. In \bibinfo{booktitle}{\emph{International Conference on Learning Representations (ICLR)}}.
\newblock


\bibitem[Dao et~al\mbox{.}(2022)]%
        {dao2022flashattention}
\bibfield{author}{\bibinfo{person}{Tri Dao}, \bibinfo{person}{Daniel~Y. Fu}, \bibinfo{person}{Stefano Ermon}, \bibinfo{person}{Atri Rudra}, {and} \bibinfo{person}{Christopher R{\'e}}.} \bibinfo{year}{2022}\natexlab{}.
\newblock \showarticletitle{Flash{A}ttention: Fast and Memory-Efficient Exact Attention with {IO}-Awareness}. In \bibinfo{booktitle}{\emph{Advances in Neural Information Processing Systems (NeurIPS)}}.
\newblock


\bibitem[De~Moura and Bj\o{}rner(2008)]%
        {z3}
\bibfield{author}{\bibinfo{person}{Leonardo De~Moura} {and} \bibinfo{person}{Nikolaj Bj\o{}rner}.} \bibinfo{year}{2008}\natexlab{}.
\newblock \showarticletitle{Z3: an efficient SMT solver}. In \bibinfo{booktitle}{\emph{Proceedings of the Theory and Practice of Software, 14th International Conference on Tools and Algorithms for the Construction and Analysis of Systems}} (Budapest, Hungary) \emph{(\bibinfo{series}{TACAS'08/ETAPS'08})}. \bibinfo{publisher}{Springer-Verlag}, \bibinfo{address}{Berlin, Heidelberg}, \bibinfo{pages}{337–340}.
\newblock
\showISBNx{3540787992}


\bibitem[Fedyukovich et~al\mbox{.}(2019)]%
        {arrayinv}
\bibfield{author}{\bibinfo{person}{Grigory Fedyukovich}, \bibinfo{person}{Sumanth Prabhu}, \bibinfo{person}{Kumar Madhukar}, {and} \bibinfo{person}{Aarti Gupta}.} \bibinfo{year}{2019}\natexlab{}.
\newblock \showarticletitle{Quantified Invariants via Syntax-Guided Synthesis}. In \bibinfo{booktitle}{\emph{Computer Aided Verification}}, \bibfield{editor}{\bibinfo{person}{Isil Dillig} {and} \bibinfo{person}{Serdar Tasiran}} (Eds.). \bibinfo{publisher}{Springer International Publishing}, \bibinfo{address}{Cham}, \bibinfo{pages}{259--277}.
\newblock
\showISBNx{978-3-030-25540-4}


\bibitem[Grosser et~al\mbox{.}(2013)]%
        {tiling}
\bibfield{author}{\bibinfo{person}{Tobias Grosser}, \bibinfo{person}{Albert Cohen}, \bibinfo{person}{Paul H.~J. Kelly}, \bibinfo{person}{J. Ramanujam}, \bibinfo{person}{P. Sadayappan}, {and} \bibinfo{person}{Sven Verdoolaege}.} \bibinfo{year}{2013}\natexlab{}.
\newblock \showarticletitle{Split tiling for GPUs: automatic parallelization using trapezoidal tiles} \emph{(\bibinfo{series}{GPGPU-6})}. \bibinfo{publisher}{Association for Computing Machinery}, \bibinfo{address}{New York, NY, USA}, \bibinfo{pages}{24–31}.
\newblock
\showISBNx{9781450320177}
\urldef\tempurl%
\url{https://doi.org/10.1145/2458523.2458526}
\showDOI{\tempurl}


\bibitem[Hoare(1969)]%
        {hoare}
\bibfield{author}{\bibinfo{person}{C.~A.~R. Hoare}.} \bibinfo{year}{1969}\natexlab{}.
\newblock \showarticletitle{An axiomatic basis for computer programming}.
\newblock \bibinfo{journal}{\emph{Commun. ACM}} \bibinfo{volume}{12}, \bibinfo{number}{10} (\bibinfo{date}{Oct.} \bibinfo{year}{1969}), \bibinfo{pages}{576–580}.
\newblock
\showISSN{0001-0782}
\urldef\tempurl%
\url{https://doi.org/10.1145/363235.363259}
\showDOI{\tempurl}


\bibitem[Hsu et~al\mbox{.}(2024)]%
        {hsu2024ligerkernelefficienttriton}
\bibfield{author}{\bibinfo{person}{Pin-Lun Hsu}, \bibinfo{person}{Yun Dai}, \bibinfo{person}{Vignesh Kothapalli}, \bibinfo{person}{Qingquan Song}, \bibinfo{person}{Shao Tang}, \bibinfo{person}{Siyu Zhu}, \bibinfo{person}{Steven Shimizu}, \bibinfo{person}{Shivam Sahni}, \bibinfo{person}{Haowen Ning}, {and} \bibinfo{person}{Yanning Chen}.} \bibinfo{year}{2024}\natexlab{}.
\newblock \showarticletitle{Liger Kernel: Efficient Triton Kernels for LLM Training}.
\newblock \bibinfo{journal}{\emph{arXiv preprint arXiv:2410.10989}} (\bibinfo{year}{2024}).
\newblock
\showeprint[arxiv]{2410.10989}~[cs.LG]
\urldef\tempurl%
\url{https://arxiv.org/abs/2410.10989}
\showURL{%
\tempurl}


\bibitem[Ish-Shalom et~al\mbox{.}(2020)]%
        {array2}
\bibfield{author}{\bibinfo{person}{Oren Ish-Shalom}, \bibinfo{person}{Shachar Itzhaky}, \bibinfo{person}{Noam Rinetzky}, {and} \bibinfo{person}{Sharon Shoham}.} \bibinfo{year}{2020}\natexlab{}.
\newblock \showarticletitle{Putting the Squeeze on Array Programs: Loop Verification via Inductive Rank Reduction}. In \bibinfo{booktitle}{\emph{Verification, Model Checking, and Abstract Interpretation: 21st International Conference, VMCAI 2020, New Orleans, LA, USA, January 16–21, 2020, Proceedings}} (New Orleans, LA, USA). \bibinfo{publisher}{Springer-Verlag}, \bibinfo{address}{Berlin, Heidelberg}, \bibinfo{pages}{112–135}.
\newblock
\showISBNx{978-3-030-39321-2}
\urldef\tempurl%
\url{https://doi.org/10.1007/978-3-030-39322-9_6}
\showDOI{\tempurl}


\bibitem[Kamil et~al\mbox{.}(2016)]%
        {stencil}
\bibfield{author}{\bibinfo{person}{Shoaib Kamil}, \bibinfo{person}{Alvin Cheung}, \bibinfo{person}{Shachar Itzhaky}, {and} \bibinfo{person}{Armando Solar-Lezama}.} \bibinfo{year}{2016}\natexlab{}.
\newblock \showarticletitle{Verified lifting of stencil computations}.
\newblock \bibinfo{journal}{\emph{SIGPLAN Not.}} \bibinfo{volume}{51}, \bibinfo{number}{6} (\bibinfo{date}{June} \bibinfo{year}{2016}), \bibinfo{pages}{711–726}.
\newblock
\showISSN{0362-1340}
\urldef\tempurl%
\url{https://doi.org/10.1145/2980983.2908117}
\showDOI{\tempurl}


\bibitem[Kloberdanz et~al\mbox{.}(2022)]%
        {deepstability}
\bibfield{author}{\bibinfo{person}{Eliska Kloberdanz}, \bibinfo{person}{Kyle~G. Kloberdanz}, {and} \bibinfo{person}{Wei Le}.} \bibinfo{year}{2022}\natexlab{}.
\newblock \showarticletitle{DeepStability: a study of unstable numerical methods and their solutions in deep learning}. In \bibinfo{booktitle}{\emph{Proceedings of the 44th International Conference on Software Engineering}} (Pittsburgh, Pennsylvania) \emph{(\bibinfo{series}{ICSE '22})}. \bibinfo{publisher}{Association for Computing Machinery}, \bibinfo{address}{New York, NY, USA}, \bibinfo{pages}{586–597}.
\newblock
\showISBNx{9781450392211}
\urldef\tempurl%
\url{https://doi.org/10.1145/3510003.3510095}
\showDOI{\tempurl}


\bibitem[Larraz et~al\mbox{.}(2013)]%
        {array1}
\bibfield{author}{\bibinfo{person}{Daniel Larraz}, \bibinfo{person}{Enric Rodr{\'i}guez-Carbonell}, {and} \bibinfo{person}{Albert Rubio}.} \bibinfo{year}{2013}\natexlab{}.
\newblock \showarticletitle{SMT-Based Array Invariant Generation}. In \bibinfo{booktitle}{\emph{Verification, Model Checking, and Abstract Interpretation}}, \bibfield{editor}{\bibinfo{person}{Roberto Giacobazzi}, \bibinfo{person}{Josh Berdine}, {and} \bibinfo{person}{Isabella Mastroeni}} (Eds.). \bibinfo{publisher}{Springer Berlin Heidelberg}, \bibinfo{address}{Berlin, Heidelberg}, \bibinfo{pages}{169--188}.
\newblock
\showISBNx{978-3-642-35873-9}


\bibitem[Lattner et~al\mbox{.}(2021)]%
        {mlir}
\bibfield{author}{\bibinfo{person}{Chris Lattner}, \bibinfo{person}{Mehdi Amini}, \bibinfo{person}{Uday Bondhugula}, \bibinfo{person}{Albert Cohen}, \bibinfo{person}{Andy Davis}, \bibinfo{person}{Jacques Pienaar}, \bibinfo{person}{River Riddle}, \bibinfo{person}{Tatiana Shpeisman}, \bibinfo{person}{Nicolas Vasilache}, {and} \bibinfo{person}{Oleksandr Zinenko}.} \bibinfo{year}{2021}\natexlab{}.
\newblock \showarticletitle{{{MLIR}}: Scaling Compiler Infrastructure for Domain Specific Computation}. In \bibinfo{booktitle}{\emph{2021 {{IEEE/ACM}} International Symposium on Code Generation and Optimization (CGO)}}. \bibinfo{pages}{2--14}.
\newblock
\urldef\tempurl%
\url{https://doi.org/10.1109/CGO51591.2021.9370308}
\showDOI{\tempurl}


\bibitem[Li and Gopalakrishnan(2010)]%
        {gpu1}
\bibfield{author}{\bibinfo{person}{Guodong Li} {and} \bibinfo{person}{Ganesh Gopalakrishnan}.} \bibinfo{year}{2010}\natexlab{}.
\newblock \showarticletitle{Scalable SMT-based verification of GPU kernel functions}. In \bibinfo{booktitle}{\emph{Proceedings of the Eighteenth ACM SIGSOFT International Symposium on Foundations of Software Engineering}} (Santa Fe, New Mexico, USA) \emph{(\bibinfo{series}{FSE '10})}. \bibinfo{publisher}{Association for Computing Machinery}, \bibinfo{address}{New York, NY, USA}, \bibinfo{pages}{187–196}.
\newblock
\showISBNx{9781605587912}
\urldef\tempurl%
\url{https://doi.org/10.1145/1882291.1882320}
\showDOI{\tempurl}


\bibitem[Lieberman(2000)]%
        {pbe}
\bibfield{author}{\bibinfo{person}{Henry Lieberman}.} \bibinfo{year}{2000}\natexlab{}.
\newblock \showarticletitle{Programming by example (introduction)}.
\newblock \bibinfo{journal}{\emph{Commun. ACM}} \bibinfo{volume}{43}, \bibinfo{number}{3} (\bibinfo{date}{March} \bibinfo{year}{2000}), \bibinfo{pages}{72–74}.
\newblock
\showISSN{0001-0782}
\urldef\tempurl%
\url{https://doi.org/10.1145/330534.330543}
\showDOI{\tempurl}


\bibitem[Liew et~al\mbox{.}(2024)]%
        {gpu2}
\bibfield{author}{\bibinfo{person}{Dennis Liew}, \bibinfo{person}{Tiago Cogumbreiro}, {and} \bibinfo{person}{Julien Lange}.} \bibinfo{year}{2024}\natexlab{}.
\newblock \showarticletitle{Sound and Partially-Complete Static Analysis of Data-Races in GPU Programs}.
\newblock \bibinfo{journal}{\emph{Proc. ACM Program. Lang.}} \bibinfo{volume}{8}, \bibinfo{number}{OOPSLA2}, Article \bibinfo{articleno}{357} (\bibinfo{date}{Oct.} \bibinfo{year}{2024}), \bibinfo{numpages}{28}~pages.
\newblock
\urldef\tempurl%
\url{https://doi.org/10.1145/3689797}
\showDOI{\tempurl}


\bibitem[Otter et~al\mbox{.}(2021)]%
        {9075398}
\bibfield{author}{\bibinfo{person}{Daniel~W. Otter}, \bibinfo{person}{Julian~R. Medina}, {and} \bibinfo{person}{Jugal~K. Kalita}.} \bibinfo{year}{2021}\natexlab{}.
\newblock \showarticletitle{A Survey of the Usages of Deep Learning for Natural Language Processing}.
\newblock \bibinfo{journal}{\emph{IEEE Transactions on Neural Networks and Learning Systems}} \bibinfo{volume}{32}, \bibinfo{number}{2} (\bibinfo{year}{2021}), \bibinfo{pages}{604--624}.
\newblock
\urldef\tempurl%
\url{https://doi.org/10.1109/TNNLS.2020.2979670}
\showDOI{\tempurl}


\bibitem[Paszke et~al\mbox{.}(2019)]%
        {pytorch}
\bibfield{author}{\bibinfo{person}{Adam Paszke}, \bibinfo{person}{Sam Gross}, \bibinfo{person}{Francisco Massa}, \bibinfo{person}{Adam Lerer}, \bibinfo{person}{James Bradbury}, \bibinfo{person}{Gregory Chanan}, \bibinfo{person}{Trevor Killeen}, \bibinfo{person}{Zeming Lin}, \bibinfo{person}{Natalia Gimelshein}, \bibinfo{person}{Luca Antiga}, \bibinfo{person}{Alban Desmaison}, \bibinfo{person}{Andreas K\"{o}pf}, \bibinfo{person}{Edward Yang}, \bibinfo{person}{Zach DeVito}, \bibinfo{person}{Martin Raison}, \bibinfo{person}{Alykhan Tejani}, \bibinfo{person}{Sasank Chilamkurthy}, \bibinfo{person}{Benoit Steiner}, \bibinfo{person}{Lu Fang}, \bibinfo{person}{Junjie Bai}, {and} \bibinfo{person}{Soumith Chintala}.} \bibinfo{year}{2019}\natexlab{}.
\newblock \bibinfo{booktitle}{\emph{PyTorch: an imperative style, high-performance deep learning library}}.
\newblock \bibinfo{publisher}{Curran Associates Inc.}, \bibinfo{address}{Red Hook, NY, USA}.
\newblock


\bibitem[Qiu et~al\mbox{.}(2024)]%
        {qiu2024tenspiler}
\bibfield{author}{\bibinfo{person}{Jie Qiu}, \bibinfo{person}{Colin Cai}, \bibinfo{person}{Sahil Bhatia}, \bibinfo{person}{Niranjan Hasabnis}, \bibinfo{person}{Sanjit~A. Seshia}, {and} \bibinfo{person}{Alvin Cheung}.} \bibinfo{year}{2024}\natexlab{}.
\newblock \showarticletitle{{Tenspiler: A Verified-Lifting-Based Compiler for Tensor Operations}}. In \bibinfo{booktitle}{\emph{38th European Conference on Object-Oriented Programming (ECOOP 2024)}} \emph{(\bibinfo{series}{Leibniz International Proceedings in Informatics (LIPIcs)}, Vol.~\bibinfo{volume}{313})}, \bibfield{editor}{\bibinfo{person}{Jonathan Aldrich} {and} \bibinfo{person}{Guido Salvaneschi}} (Eds.). \bibinfo{publisher}{Schloss Dagstuhl -- Leibniz-Zentrum f{\"u}r Informatik}, \bibinfo{address}{Dagstuhl, Germany}, \bibinfo{pages}{32:1--32:28}.
\newblock
\showISBNx{978-3-95977-341-6}
\showISSN{1868-8969}
\urldef\tempurl%
\url{https://doi.org/10.4230/LIPIcs.ECOOP.2024.32}
\showDOI{\tempurl}


\bibitem[Sharma et~al\mbox{.}(2015)]%
        {gpu3}
\bibfield{author}{\bibinfo{person}{Rahul Sharma}, \bibinfo{person}{Michael Bauer}, {and} \bibinfo{person}{Alex Aiken}.} \bibinfo{year}{2015}\natexlab{}.
\newblock \showarticletitle{Verification of producer-consumer synchronization in GPU programs}. In \bibinfo{booktitle}{\emph{Proceedings of the 36th ACM SIGPLAN Conference on Programming Language Design and Implementation}} (Portland, OR, USA) \emph{(\bibinfo{series}{PLDI '15})}. \bibinfo{publisher}{Association for Computing Machinery}, \bibinfo{address}{New York, NY, USA}, \bibinfo{pages}{88–98}.
\newblock
\showISBNx{9781450334686}
\urldef\tempurl%
\url{https://doi.org/10.1145/2737924.2737962}
\showDOI{\tempurl}


\bibitem[Shi et~al\mbox{.}(2022)]%
        {tf-coder}
\bibfield{author}{\bibinfo{person}{Kensen Shi}, \bibinfo{person}{David Bieber}, {and} \bibinfo{person}{Rishabh Singh}.} \bibinfo{year}{2022}\natexlab{}.
\newblock \showarticletitle{TF-Coder: Program Synthesis for Tensor Manipulations}.
\newblock \bibinfo{journal}{\emph{ACM Trans. Program. Lang. Syst.}} \bibinfo{volume}{44}, \bibinfo{number}{2}, Article \bibinfo{articleno}{10} (\bibinfo{date}{May} \bibinfo{year}{2022}), \bibinfo{numpages}{36}~pages.
\newblock
\showISSN{0164-0925}
\urldef\tempurl%
\url{https://doi.org/10.1145/3517034}
\showDOI{\tempurl}


\bibitem[Tillet et~al\mbox{.}(2019)]%
        {triton}
\bibfield{author}{\bibinfo{person}{Philippe Tillet}, \bibinfo{person}{H.~T. Kung}, {and} \bibinfo{person}{David Cox}.} \bibinfo{year}{2019}\natexlab{}.
\newblock \showarticletitle{Triton: an intermediate language and compiler for tiled neural network computations}. In \bibinfo{booktitle}{\emph{Proceedings of the 3rd ACM SIGPLAN International Workshop on Machine Learning and Programming Languages}} (Phoenix, AZ, USA) \emph{(\bibinfo{series}{MAPL 2019})}. \bibinfo{publisher}{Association for Computing Machinery}, \bibinfo{address}{New York, NY, USA}, \bibinfo{pages}{10–19}.
\newblock
\showISBNx{9781450367196}
\urldef\tempurl%
\url{https://doi.org/10.1145/3315508.3329973}
\showDOI{\tempurl}


\bibitem[Wang et~al\mbox{.}(2010)]%
        {kernel_fusion}
\bibfield{author}{\bibinfo{person}{Guibin Wang}, \bibinfo{person}{YiSong Lin}, {and} \bibinfo{person}{Wei Yi}.} \bibinfo{year}{2010}\natexlab{}.
\newblock \showarticletitle{Kernel Fusion: An Effective Method for Better Power Efficiency on Multithreaded GPU}. In \bibinfo{booktitle}{\emph{Proceedings of the 2010 IEEE/ACM Int'l Conference on Green Computing and Communications \& Int'l Conference on Cyber, Physical and Social Computing}} \emph{(\bibinfo{series}{GREENCOM-CPSCOM '10})}. \bibinfo{publisher}{IEEE Computer Society}, \bibinfo{address}{USA}, \bibinfo{pages}{344–350}.
\newblock
\showISBNx{9780769543314}
\urldef\tempurl%
\url{https://doi.org/10.1109/GreenCom-CPSCom.2010.102}
\showDOI{\tempurl}


\bibitem[Willsey et~al\mbox{.}(2021)]%
        {egg}
\bibfield{author}{\bibinfo{person}{Max Willsey}, \bibinfo{person}{Chandrakana Nandi}, \bibinfo{person}{Yisu~Remy Wang}, \bibinfo{person}{Oliver Flatt}, \bibinfo{person}{Zachary Tatlock}, {and} \bibinfo{person}{Pavel Panchekha}.} \bibinfo{year}{2021}\natexlab{}.
\newblock \showarticletitle{egg: Fast and extensible equality saturation}.
\newblock \bibinfo{journal}{\emph{Proc. ACM Program. Lang.}} \bibinfo{volume}{5}, \bibinfo{number}{POPL}, Article \bibinfo{articleno}{23} (\bibinfo{date}{Jan.} \bibinfo{year}{2021}), \bibinfo{numpages}{29}~pages.
\newblock
\urldef\tempurl%
\url{https://doi.org/10.1145/3434304}
\showDOI{\tempurl}


\bibitem[Yang et~al\mbox{.}(2022)]%
        {10.1145/3505243}
\bibfield{author}{\bibinfo{person}{Yanming Yang}, \bibinfo{person}{Xin Xia}, \bibinfo{person}{David Lo}, {and} \bibinfo{person}{John Grundy}.} \bibinfo{year}{2022}\natexlab{}.
\newblock \showarticletitle{A Survey on Deep Learning for Software Engineering}.
\newblock \bibinfo{journal}{\emph{ACM Comput. Surv.}} \bibinfo{volume}{54}, \bibinfo{number}{10s}, Article \bibinfo{articleno}{206} (\bibinfo{date}{Sept.} \bibinfo{year}{2022}), \bibinfo{numpages}{73}~pages.
\newblock
\showISSN{0360-0300}
\urldef\tempurl%
\url{https://doi.org/10.1145/3505243}
\showDOI{\tempurl}


\bibitem[Zhang and Sennrich(2019)]%
        {rmsnorm}
\bibfield{author}{\bibinfo{person}{Biao Zhang} {and} \bibinfo{person}{Rico Sennrich}.} \bibinfo{year}{2019}\natexlab{}.
\newblock \bibinfo{booktitle}{\emph{Root mean square layer normalization}}.
\newblock \bibinfo{publisher}{Curran Associates Inc.}, \bibinfo{address}{Red Hook, NY, USA}.
\newblock


\bibitem[Zhang et~al\mbox{.}(2021a)]%
        {duo}
\bibfield{author}{\bibinfo{person}{Xufan Zhang}, \bibinfo{person}{Jiawei Liu}, \bibinfo{person}{Ning Sun}, \bibinfo{person}{Chunrong Fang}, \bibinfo{person}{Jia Liu}, \bibinfo{person}{Jiang Wang}, \bibinfo{person}{Dong Chai}, {and} \bibinfo{person}{Zhenyu Chen}.} \bibinfo{year}{2021}\natexlab{a}.
\newblock \showarticletitle{Duo: Differential Fuzzing for Deep Learning Operators}.
\newblock \bibinfo{journal}{\emph{IEEE Transactions on Reliability}} \bibinfo{volume}{70}, \bibinfo{number}{4} (\bibinfo{year}{2021}), \bibinfo{pages}{1671--1685}.
\newblock
\urldef\tempurl%
\url{https://doi.org/10.1109/TR.2021.3107165}
\showDOI{\tempurl}


\bibitem[Zhang et~al\mbox{.}(2021b)]%
        {predoo}
\bibfield{author}{\bibinfo{person}{Xufan Zhang}, \bibinfo{person}{Ning Sun}, \bibinfo{person}{Chunrong Fang}, \bibinfo{person}{Jiawei Liu}, \bibinfo{person}{Jia Liu}, \bibinfo{person}{Dong Chai}, \bibinfo{person}{Jiang Wang}, {and} \bibinfo{person}{Zhenyu Chen}.} \bibinfo{year}{2021}\natexlab{b}.
\newblock \showarticletitle{Predoo: precision testing of deep learning operators}. In \bibinfo{booktitle}{\emph{Proceedings of the 30th ACM SIGSOFT International Symposium on Software Testing and Analysis}} (Virtual, Denmark) \emph{(\bibinfo{series}{ISSTA 2021})}. \bibinfo{publisher}{Association for Computing Machinery}, \bibinfo{address}{New York, NY, USA}, \bibinfo{pages}{400–412}.
\newblock
\showISBNx{9781450384599}
\urldef\tempurl%
\url{https://doi.org/10.1145/3460319.3464843}
\showDOI{\tempurl}


\bibitem[Zhang et~al\mbox{.}(2023)]%
        {egglog}
\bibfield{author}{\bibinfo{person}{Yihong Zhang}, \bibinfo{person}{Yisu~Remy Wang}, \bibinfo{person}{Oliver Flatt}, \bibinfo{person}{David Cao}, \bibinfo{person}{Philip Zucker}, \bibinfo{person}{Eli Rosenthal}, \bibinfo{person}{Zachary Tatlock}, {and} \bibinfo{person}{Max Willsey}.} \bibinfo{year}{2023}\natexlab{}.
\newblock \showarticletitle{Better Together: Unifying Datalog and Equality Saturation}.
\newblock \bibinfo{journal}{\emph{Proc. ACM Program. Lang.}} \bibinfo{volume}{7}, \bibinfo{number}{PLDI}, Article \bibinfo{articleno}{125} (\bibinfo{date}{June} \bibinfo{year}{2023}), \bibinfo{numpages}{25}~pages.
\newblock
\urldef\tempurl%
\url{https://doi.org/10.1145/3591239}
\showDOI{\tempurl}


\end{thebibliography}

\end{document}